\documentclass[preprint,sort, compress,12pt,3p]{elsarticle}

\UseRawInputEncoding
\usepackage{moreverb}
\usepackage{amsmath}
\usepackage{algorithmic}
\usepackage{graphics}
\usepackage{graphicx}
\usepackage{subfigure}
\usepackage{enumitem}
\usepackage{color,soul}
\usepackage{framed}
\usepackage{wrapfig}
\usepackage{bm}
\usepackage{mathrsfs}
\usepackage{amsfonts}
\usepackage{multirow}
\usepackage{longtable}
\usepackage{hyperref}
\usepackage{paralist}
\usepackage{indentfirst}
\usepackage{relsize}
\usepackage{extarrows}
\usepackage{lineno,xcolor}
\usepackage{upgreek}
\usepackage{bm}
\usepackage{xcolor}
\usepackage{booktabs}
\usepackage[flushleft]{threeparttable}
\usepackage[linesnumbered,lined,ruled]{algorithm2e}
\usepackage{color, colortbl}
\usepackage{dcolumn}
\usepackage{tocloft}
\definecolor{Gray}{gray}{0.9}
\usepackage{float}

\usepackage{geometry}
\graphicspath{ {./Figure/} }

\hypersetup{
bookmarks=true,
bookmarksopen=true,
bookmarksnumbered=true,
unicode=false,
pdftoolbar=true,
pdfmenubar=true,
pdffitwindow=false,
pdfstartview={FitH},
pdftitle={My title},
pdfauthor={Author},
pdfsubject={Subject},
pdfcreator={Creator},
pdfproducer={Producer},
pdfkeywords={keywords},
pdfnewwindow=true,
colorlinks=true,
linkcolor=blue,
citecolor=blue,
filecolor=magenta,
urlcolor=blue
}

\begin{document}

\title{\Large MC-GRU:a Multi-Channel GRU network for generalized nonlinear structural response prediction across structures}

\author[SEU]{Shan He}
\author[SEU,SEU2,SEU3]{Ruiyang Zhang\corref{cor}}
\ead{ryzhang@seu.edu.cn}

\cortext[cor]{Corresponding author. Tel: +86 15996466907}

\address[SEU]{Key Laboratory of Concrete and Prestressed Concrete Structures of the Ministry of Education, Southeast University, Nanjing, China}
\address[SEU2]{National and Local Joint Engineering Research Center for Intelligent Construction and Maintenance, Southeast University, Nanjing, China}
\address[SEU3]{School of Civil Engineering, Southeast University, Nanjing, China}

\begin{abstract}
    \small
    
    Accurate prediction of seismic responses and quantification of structural damage are critical in civil engineering. Traditional approaches such as finite element analysis could lack computational efficiency, especially for complex structural systems under extreme hazards. Recently, artificial intelligence has provided an alternative to efficiently model highly nonlinear behaviors. However, existing models face challenges in generalizing across diverse structural systems. This paper proposes a novel multi-channel gated recurrent unit (MC-GRU) network aimed at achieving generalized nonlinear structural response prediction for varying structures. The key concept lies in the integration of a multi-channel input mechanism to GRU with an extra input of structural information to the candidate hidden state, which enables the network to learn the dynamic characteristics of diverse structures and thus empower the generalizability and adaptiveness to unseen structures. The performance of the proposed MC-GRU is validated through a series of case studies, including a single-degree-of-freedom  linear system, a hysteretic Bouc-Wen system, and a nonlinear reinforced concrete column from experimental testing. Results indicate that the proposed MC-GRU overcomes the major generalizability issues of existing methods, with capability of accurately inferring seismic responses of varying structures. Additionally, it demonstrates enhanced capabilities in representing nonlinear structural dynamics compared to traditional models such as GRU and LSTM.

\end{abstract}


\maketitle

\section{Introduction}\label{sec1}
In recent years, the design of structures to resist natural disasters, particularly earthquakes, has become a priority in civil engineering \cite{1fema2013hazus}. Earthquakes, as one of the most common natural disasters, have historically caused significant damage to structures, resulting in considerable economic losses and casualties \cite{2kazama2012damage,3di2019seismic}. Therefore, in the context of structural seismic design, especially performance-based earthquake engineering (PBEE) \cite{4roy2021integral,5damikoukas2021direct,6HUANG20212396,ES-zhou;pbee,ES-zhoupbee2}, accurately predicting seismic responses and quantifying damage levels are essential. Numerical simulation methods based on physical equations, such as finite element method (FEM), are currently the most commonly used tools for simulating and analyzing the behavior and response of buildings under seismic actions\cite{7moaveni2009uncertainty,8GIRARDI2021107372,9sun2015hybrid,10doi:10.1193/1.2219487, ES-zhoufem}. FEM divides the structure into numerous small elements and nodes, forming a mathematical model to simulate the building's behavior and response to seismic activities. By applying suitable material models, such as elastic or plastic models, FEM forecasts the deformation and damage of buildings during earthquakes. However, the precision of FEM depends on accurate assumptions of material behavior and boundary conditions, and it demands considerable computational resources and time.

Data-driven response prediction methods have increasingly garnered the attention of engineers. These approaches create a mapping relationship between seismic activities and structural responses by leveraging extensive historical data. Various machine learning techniques, such as Support Vector Regression (SVR)\cite{11trafalis2000support} and basic Multilayer Perceptrons (MLP) \cite{12abbas2020prediction,13de2009prediction,14nikose2019dynamic,15oh2020neural,16lu2005underground}, have successfully predicted the nonlinear response of structures, demonstrating notable computational efficiency. Recently, with advancements in hardware, deep learning methods have been introduced to response prediction tasks. Compared to traditional machine learning, deep learning methods can develop more complex network structures, thus more effectively capturing the nonlinear behavior of structures. Convolutional Neural Networks (CNNs) \cite{17CNN:10.5555/303568.303704} excel in extracting spatial features from data, effectively capturing latent characteristics of seismic sequences through sliding convolutional kernels. CNNs have been successfully utilized for predicting seismic responses \cite{18wu2019deep,19zhang2020physics,20sun2017data,21kim2019response,ES-cnn,ES-cnn2} and detecting structural damage \cite{23truong2022effective,24truong2022joint}. However, the sliding window approach of CNNs struggles with capturing long-range dependencies in sequences. Recurrent Neural Networks (RNNs) address this limitation \cite{25RNN:10.1162/neco.1989.1.2.270}, as they are specifically designed for sequence data and can retain previous information through internal loops. Among RNNs, Long Short-Term Memory (LSTM) networks \cite{26LSTM:6795963} have demonstrated superior performance. Zhang et al. \cite{27zhang2019deep} were the first to use LSTM networks for predicting the nonlinear responses of complex dynamic systems. Xu and Lu \cite{Lu2021real} propose a real-time regional seismic damage assessment framework using a LSTM neural network, which establishes high-performance mapping between ground motions and structural damage through region-specific models. Validated on Tsinghua University campus buildings, the framework achieves high prediction accuracy and real-time capability, offering a novel solution for rapid post-earthquake response at regional scales. 
Researchers have also proposed various physics-guided methods based on deep learning to enhance the predictive performance of these networks, achieving notable improvements in accuracy \cite{28ZHANG2020113226,29WANG2023115576,ES-py}. Lu et al. \cite{lu2022PHY} introduced a Physics Estimator into the model training process to evaluate the physical behavior of generated designs, enhancing the objectiveness of training. Fei and Lu et al. \cite{fei2024hybrid} proposed a hybrid surrogate model integrating physics-based and data-driven approaches to efficiently estimate building seismic responses. They enhanced dataset diversity through data augmentation, introduced a task decomposition strategy combining a graph neural network (GNN) \cite{GNN} with a flexural-shear model, and optimized the GNN’s output layer and loss function to improve accuracy, achieving significantly faster and more precise inter-story drift ratio (IDR) estimations than existing methods.
The attention mechanism, initially proposed for machine translation tasks, is a crucial technique for enhancing model performance in sequence data processing and has also been applied to nonlinear response prediction. For instance, Liao et al. \cite{30liao2023attention} incorporated the attention mechanism into LSTM networks, improving the capture of nonlinear behavior in bridges during earthquakes. Wang \cite{31WANG2020113357} employed the Unroll Attention Sequence to Sequence (UA-Seq2Seq) model, successfully applying it to the hysteresis curves of components. Li et al. \cite{32:9328806} proposed a time series attention-based RNN encoder-decoder architecture for predicting structural responses under seismic excitation.

Current research primarily aims to enhance the prediction accuracy of nonlinear responses through the optimization of neural networks. However, a significant limitation of these algorithms is their lack of generality across various structures. The work of Kuo et al. \cite{KUO2024117733} attempts to address this issue by representing building structures as graphs and integrating GNN  with LSTM, enabling a model that predicts dynamic responses for various structures. Similarly, Shu et al. \cite{SHU2024110628} proposed a TransFrameNet based on Transformer \cite{Trans} architecture, which considers variations in geometric features and component sizes across different buildings by converting buildings into archetypes. However, despite the incorporation of structural features during training, these studies only exhibit limited generalizability for unseen structures out of the training set. To overcome this limitation, this paper proposes a novel multi-channel gated recurrent unit (MC-GRU) network for generalized nonlinear structural response prediction. The MC-GRU is specifically designed with a multi-channel input mechanism based on classical GRU-based gated units to capture the dynamic features of both ground motion and structural information on response prediction. This integration allows for generalization at the structural information level, while also demonstrating superior performance in predicting nonlinear structural responses across diverse unseen structures. 

The organization of the rest of this paper is as follows:
Section \ref{sec1-2} provides a detailed problem statement with emphasize on the generalizability issues of existing AI-based surrogate modeling of nonlinear responses. Section \ref{sec2} introduces the details of the multi-input mechanism and the proposed MC-GRU network. Section \ref{sec3} validates the predictive performance of MC-GRU through examples including a single-degree-of-freedom (SDOF) linear system, a SDOF nonlinear Bouc-Wen system, and a reinforced concrete column. Finally, Section \ref{sec_con} summarizes the major conclusions of this study.

\section{Problem statement}\label{sec1-2}
\begin{figure}
    \centering
    \includegraphics[width=1.0\linewidth]{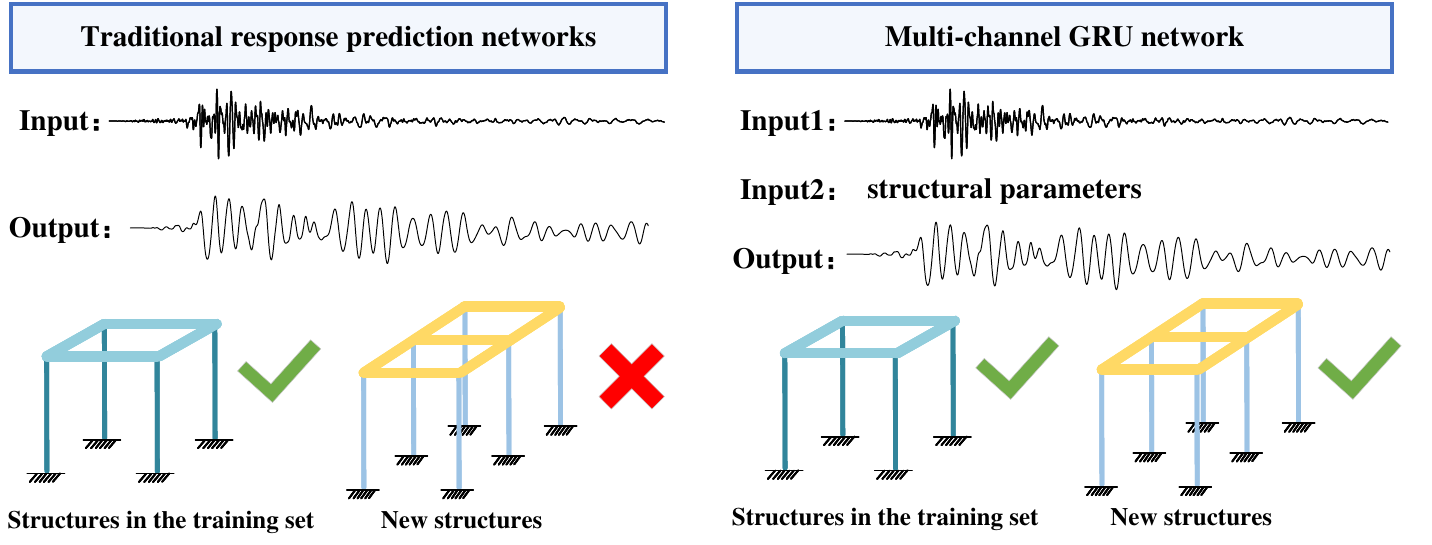}
    \caption{Comparison between the MC-GRU network and traditional response prediction networks.}
    \label{fig:probstatement}
\end{figure}

This study aims to address the critical challenge of generalization issues of AI-based surrogate model for unseen structures. The core of this limitation lies in the sequence-to-sequence (seq2seq) paradigm, which underpins most existing models. The seq2seq paradigm performs well in common sequential tasks because the mapping relationships in these tasks are typically fixed, such as in stock price prediction or machine translation. However, in structure metamodeling tasks, the mapping relationship between ground motions and structural responses represent the dynamic properties of a specific structure, and thus it becomes invalid when structural parameters change, rendering the seq2seq paradigm inadequate for cross-structural prediction tasks (illustrated in Figure \ref{fig:probstatement}). To solve this challenge, one potential solution is to consider structural information as extra input to the neural network such that it can learn dynamic characteristics of varying structures. 

For building-type structures, their dynamic behavior is typically governed by a nonlinear equation of motion. Consider an example of a nonlinear single-degree-of-freedom system for illustration:
\begin{flalign}
    \label{eq1}
&M\ddot{x}\left(t\right)+C\dot{x}\left(t\right)+F_s=-M{\ddot{x}}_g\left(t\right)&
\end{flalign}
where \(x\), \(\dot{x}\), and \(\ddot{x}\) are the vectors of structural displacement, velocity, and acceleration respectively; \(M\), \(C\), and \(F_s\) are the mass matrix, damping matrix, and restoring force; \({\ddot{x}}_g\) is the ground motion acceleration;

The damping matrix \(C\) is approximated through the Rayleigh damping model, which assumes that the structural damping is proportional to the structural mass and stiffness, say \(C=\alpha M+\beta K\) in which \(\alpha\) and \(\beta\) are the damping coefficients, and \(K\) is the structural stiffness matrix. The restoring force \(F_s\) is determined by the stiffness matrix and the hysteretic characteristics of the system.

Predicting the dynamic response of a structure involves solving the above equation of motion. In previous studies, deep learning models were mainly developed using data-driven approaches to establish the mapping between ground motion acceleration and structural responses. However, it is crucial to recognize that a structure's mass and stiffness also play significant roles in its response to specific seismic events. This study proposes a novel MC-GRU network for generalized nonlinear structural response prediction across diverse structures. Unlike traditional seq2seq models, the MC-GRU incorporates both ground motion data and structural parameters as the input, establishing a mapping from sequences and parameters to responses, as illustrated in Figure \ref{fig:probstatement}. Specifically, the proposed model can be expressed as:
\begin{flalign}
    \label{eq0}
&Y = \text{MC-GRU}(X, S; \Phi)&
\end{flalign}
where \(X\) represents the ground motion data, \(S\) denotes the structural parameters, \(Y\) is the structural response, and \(\Phi\) represents the trainable parameters of the neural network. 

\section{Multi-channel GRU network}\label{sec2}
\begin{figure}
    \centering
    \includegraphics[width=0.8\linewidth]{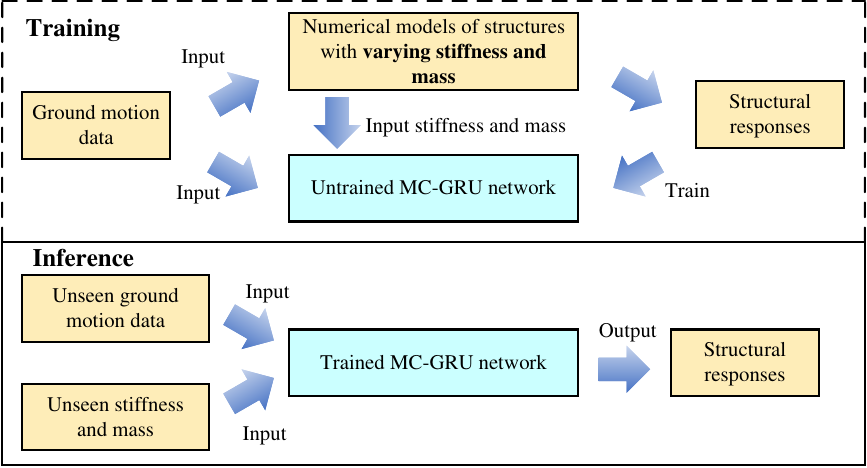}
    \caption{The flowchart of MC-GRU for generalized nonlinear structural response prediction.}
    \label{fig:flowchart}
\end{figure}

This section introduces the methodology of the proposed MC-GRU network for generalized nonlinear structural response prediction across multi-structures. A multi-input mechanism is specifically designed to consider extra input of structural information to GRU. This multi-channel architecture enables the network to concurrently learn the impacts of both ground motion and structural parameters on nonlinear responses, empowering the generalizability and adaptiveness to unseen structures. The flowchart of the proposed methodology is presented in Figure \ref{fig:flowchart}. During the training phase, ground motion data and structural information (e.g., structural stiffness and mass) are fed into the MC-GRU network for model training. The trained MC-GRU network is then used to infer the nonlinear seismic responses of diverse structures given unseen ground motion and structural parameters as input. 

\subsection{MC-GRU cell}\label{sec2.3}

The GRU network has been shown to be a promising model for predicting structural responses, possessing fewer parameters than LSTM while demonstrating superior performance in specific tasks \cite{35GRU:8053243}. To account for the structural information in the learning of the structural dynamic characteristics, a multi-input mechanism is specifically designed based on GRU. Figure \ref{fig：MC-GRU cell} presents the structure of the proposed MC-GRU cell, which incorporates the reset gate and update gate from GRU, designed with an extra input for structural information in addition to the ground motion. We denote, at time \(t\) within MC-GRU layer \(l\), the input of ground motion as \({x}_t^{\left(l\right)}\), the input of structural information as \(\boldsymbol{S}^{\left(l\right)}\), the reset gate as  \({r}_t^{\left(l\right)}\), the update gate as \({z}_t^{\left(l\right)}\), and the hidden state output as \({h}_t^{\left(l\right)}\). Then, the variables are computed as follows:

\begin{figure}
    \centering
    \includegraphics[width=1.0\linewidth]{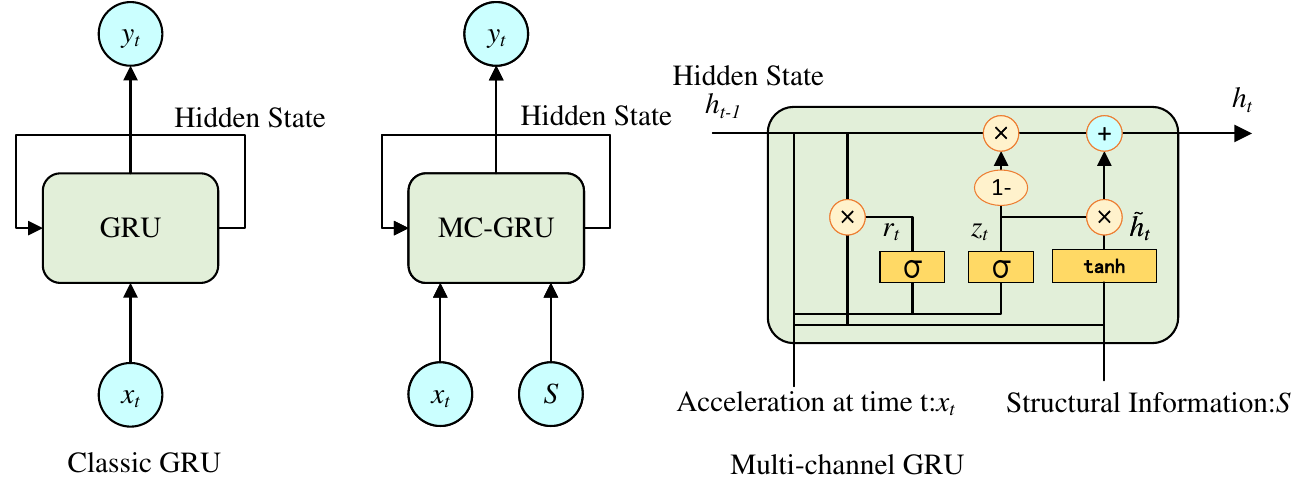}
    \caption{The architecture of MC-GRU cell.}
    \label{fig：MC-GRU cell}
\end{figure}

\begin{flalign}
    \label{eq2}
&{r}_t^{\left(l\right)}=\sigma\left({W}_{xr}{x}_t+{W}_{hr}{h}_{t-1}^{\left(l\right)}+{b}_r\right)&
\\
    \label{eq3}
&{z}_t^{\left(l\right)}=\sigma\left({W}_{xz}{x}_t+{W}_{hz}{h}_{t-1}^{\left(l\right)}+{b}_z\right)&
\\
    \label{eq4}
&\boldsymbol{S}^{\left(l\right)}={W}_{sh}Stiffness+{W}_{mh}Mass+{b}_s&
\\
    \label{eq5}
&{\widetilde{{h}}}_t^{\left(l\right)}=\tanh{\left({W}_{xh}{x}_t+{W}_{hh}\left({r}_t^{\left(l\right)}\odot{h}_{t-1}^{\left(l\right)}\right)+\boldsymbol{S}^{\left(l\right)}+{b}_h\right)}&
\\
    \label{eq6}
&{h}_t^{\left(l\right)}=\left(1-{z}_t^{\left(l\right)}\right)\odot{h}_{t-1}^{\left(l\right)}+{z}_t^{\left(l\right)}\odot{\widetilde{{h}}}_t^{\left(l\right)}&
\end{flalign}

where, \({W}_{xr}\), \({W}_{hr}\), \({W}_{xz}\), \({W}_{hz}\), \({W}_{sh}\), \({W}_{mh}\), \({W}_{xh}\) and \({W}_{hh}\), are weight matrices; \({b}_r\), \({b}_z\), \({b}_s\), \({b}_h\) are bias vectors; \(\sigma\left(\right)\) is sigmoid function; \(tanh\left(\right)\) is hyperbolic tangent function.

In the MC-GRU cell, information flow adheres to the identical pattern observed in the GRU cell \cite{35GRU:8053243}. During the calculation of the candidate hidden state \({\widetilde{{h}}}_t^{\left(l\right)}\), the impact of the structural state \({S}^{\left(l\right)}\) is integrated. Both the mass and the stiffness of the structure jointly determine the value of \({S}^{\left(l\right)}\).

\subsection{MC-GRU for predicting seismic responses}\label{sec2.4}

\begin{figure}
    \centering
    \includegraphics[width=1.0\linewidth]{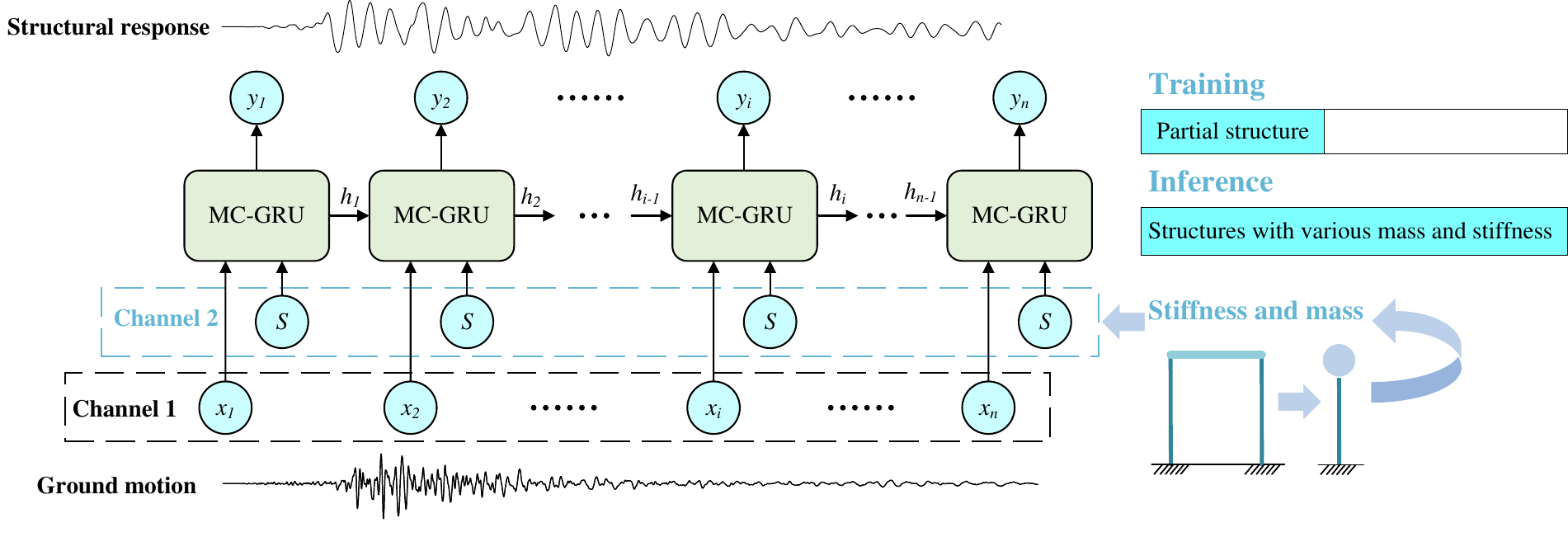}
    \caption{The architecture of the MC-GRU network.}
    \label{fig:MC-GRU network}
\end{figure}

The architecture of the MC-GRU network is illustrated in Figure \ref{fig:MC-GRU network}. The network comprises two input channels, each handling distinct types of data. \textbf{Channel 1} receives ground motion sequences, represented as a tensor with dimensions [batch\_size, sequence\_length, input\_size1], where batch\_size denotes the number of samples per training batch, and sequence\_length denotes the length of the ground motion sequence. In this study, ground motions are sampled at intervals of 0.02 seconds over a total duration of 30 seconds, resulting in sequence\_length = 1500. The input\_size denotes the dimension of the ground motions; here, unidirectional seismic intensity is utilized, yielding input\_size = 1. \textbf{Channel 2} processes structural information, represented by a tensor with the shape [batch\_size, sequence\_length, input\_size2], where input\_size2 denotes the number of structural information parameters. In this study, the structural parameters encompass mass and stiffness, thus input\_size2 = 2. The output shape of the network is [batch\_size, sequence\_length, output\_size], where output\_size denotes the dimension of displacement. In the case of a single-degree-of-freedom system, output\_size = 1. To facilitate training, all input data are normalized to range [-1,1] to improve the convergence during optimization.

The Adam optimizer was selected for model training considering its low memory requirement and suitability for non-stationary problems \cite{38kingma2014adam}. The loss function is determined as the mean square error (MSE) for regression problems. The entire training process is conducted in a Python environment using Pytorch \cite{39paszke2019pytorch}. Simulations are performed on a standard PC with one Intel Xeon Platinum 8352S CPU and three NVIDIA L40 48G GPUs. The data and codes used in this paper will be publicly available on GitHub at https://github.com/SEU-bobi/MC-GRU after the paper is published.

\section{Numerical Validation Examples}\label{sec3}
This section presents the numerical validation of the proposed MC-GRU for generalized nonlinear responses across diverse structures. Three case studies are investigated to validate the proposed approach including an SDOF linear system, a Bouc-Wen hysteresis model, and a nonlinear reinforced concrete column.
\subsection{Case 1: Single degree of freedom linear system}\label{sec3.1}
A simplified SDOF linear system is utilized to evaluate the generalization capacity of the proposed MC-GRU across various structures. The governing equation of motion of the SDOF linear system can be expressed as follows:
\begin{flalign}
    \label{eq:linearSDOF}
&m\ddot{x}\left(t\right)+c\dot{x}\left(t\right)+kx\left(t\right)=-m{\ddot{x}}_g\left(t\right)&
\end{flalign}
where \(x\), \(\dot{x}\), and \(\ddot{x}\) are the vectors of structural displacement, velocity, and acceleration respectively; \(m\), \(c\), and \(k\) are the mass, damping, and stiffness; \(\ddot{x}_g\) is the ground motion acceleration.

To generate the data for network training and testing, a total of 42 ground motions are selected from the NGA-West2 ground motion database \cite{40ancheta2014nga}. The dataset is partitioned into three parts: 20 for training, 16 for validation, and 6 for testing. To increase the diversity of earthquake samples, the amplitude of each ground motion is scaled, with peak ground accelerations (PGA) set at 0.2g, 0.4g, 0.6g, 0.8g, and 1.0g, respectively. Ultimately, the training set included 100 samples of seismic data, the validation set comprised 80 samples, and the test set contained 30 samples.

 In addition, to ensure the model's ability to generalize across various structures, 89 structures with different stiffness and mass combinations are included in the training and testing datasets. Table \ref{table1} summarizes the structural information used in  model training. The training dataset consists of 25 distinct combinations of stiffness and mass, with mass ranging from 120 kg to 240 kg and stiffness ranging from 30 kN/m to 65 kN/m. The natural frequencies of the structures range from 1.78 Hz to 3.70 Hz. Table \ref{table2} summarizes the structural information used in validation and testing. The validation and testing datasets include 64 diverse combinations of stiffness and mass, with mass ranging from 80 kg to 300 kg and stiffness ranging from 20 kN/m to 100 kN/m. The natural frequencies of the system range from 1.30 Hz to 5.63 Hz. 

Overall, the training dataset includes 2500 samples from 100 ground motion records and 25 sets of structural information, the validation dataset consists of 5120 samples from 80 ground motions and 64 structural information, and the testing dataset comprises 1920 samples from 30 ground motion records and 64 sets of structural information. Each sample has a pairwise input of ground motion and structural information and output of structural displacement responses.

\begin{table}[t!]
	\caption{Structural information in training datasets.} \vspace{6pt}
	\label{table1}
	\footnotesize
	\centering
	\setlength{\tabcolsep}{5mm}{
	\begin{tabular}{ccccccc} 
	\hline 
	No. & Stiffness(kN/m) & Mass(kg) & Natural frequency(Hz) \\
	\hline
        1   & 30              & 120      & 2.52                    \\
        2   & 30              & 140      & 2.33                    \\
        3   & 30              & 160      & 2.18                    \\
        4   & 30              & 200      & 1.95                    \\
        5   & 30              & 240      & 1.78                    \\
        6   & 35              & 120      & 2.72                    \\
        7   & 35              & 140      & 2.52                    \\
        8   & 35              & 160      & 2.35                    \\
        9   & 35              & 200      & 2.11                    \\
        10  & 35              & 240      & 1.92                    \\
        11  & 40              & 120      & 2.91                    \\
        12  & 40              & 140      & 2.69                    \\
        13  & 40              & 160      & 2.52                    \\
        14  & 40              & 200      & 2.25                    \\
        15  & 40              & 240      & 2.05                    \\
        16  & 55              & 120      & 3.41                    \\
        17  & 55              & 140      & 3.15                    \\
        18  & 55              & 160      & 2.95                    \\
        19  & 55              & 200      & 2.63                    \\
        20  & 55              & 240      & 2.41                    \\
        21  & 65              & 120      & 3.70                    \\
        22  & 65              & 140      & 3.43                    \\
        23  & 65              & 160      & 3.21                    \\
        24  & 65              & 200      & 2.87                    \\
        25  & 65              & 240      & 2.62                    \\ 
	\hline
	\end{tabular}}
\end{table}

\begin{table}[t!]
	\caption{Structural information in validation and testing datasets.} \vspace{6pt}
	\label{table2}
	\footnotesize
	\centering
	\setlength{\tabcolsep}{2mm}{
	\begin{tabular}{cccccccccc} 
	\hline 
	No. & \begin{tabular}[c]{@{}c@{}}Stiffness \\ (kN/m)\end{tabular} & \begin{tabular}[c]{@{}c@{}}Mass \\ (kg)\end{tabular} & \begin{tabular}[c]{@{}c@{}} Natural frequency\\(Hz)\end{tabular} &No. & \begin{tabular}[c]{@{}c@{}}Stiffness \\ (kN/m)\end{tabular} & \begin{tabular}[c]{@{}c@{}}Mass \\ (kg)\end{tabular} & \begin{tabular}[c]{@{}c@{}} Natural frequency\\(Hz)\end{tabular} \\
	\hline
        1   & 20              & 80       & 2.52                    & 33  & 60              & 80       & 4.36                    \\
2   & 20              & 100      & 2.25                    & 34  & 60              & 100      & 3.90                    \\
3   & 20              & 150      & 1.84                    & 35  & 60              & 150      & 3.18                    \\
4   & 20              & 180      & 1.68                    & 36  & 60              & 180      & 2.91                    \\
5   & 20              & 220      & 1.52                    & 37  & 60              & 220      & 2.63                    \\
6   & 20              & 260      & 1.40                    & 38  & 60              & 260      & 2.42                    \\
7   & 20              & 280      & 1.35                    & 39  & 60              & 280      & 2.33                    \\
8   & 20              & 300      & 1.30                    & 40  & 60              & 300      & 2.25                    \\
9   & 25              & 80       & 2.81                    & 41  & 75              & 80       & 4.87                    \\
10  & 25              & 100      & 2.51                    & 42  & 75              & 100      & 4.36                    \\
11  & 25              & 150      & 2.05                    & 43  & 75              & 150      & 3.56                    \\
12  & 25              & 180      & 1.88                    & 44  & 75              & 180      & 3.25                    \\
13  & 25              & 220      & 1.70                    & 45  & 75              & 220      & 2.94                    \\
14  & 25              & 260      & 1.56                    & 46  & 75              & 260      & 2.70                    \\
15  & 25              & 280      & 1.50                    & 47  & 75              & 280      & 2.60                    \\
16  & 25              & 300      & 1.45                    & 48  & 75              & 300      & 2.52                    \\
17  & 38              & 80       & 3.47                    & 49  & 90              & 80       & 5.34                    \\
18  & 38              & 100      & 3.10                    & 50  & 90              & 100      & 4.78                    \\
19  & 38              & 150      & 2.53                    & 51  & 90              & 150      & 3.90                    \\
20  & 38              & 180      & 2.31                    & 52  & 90              & 180      & 3.56                    \\
21  & 38              & 220      & 2.09                    & 53  & 90              & 220      & 3.21                    \\
22  & 38              & 260      & 1.92                    & 54  & 90              & 260      & 2.96                    \\
23  & 38              & 280      & 1.85                    & 55  & 90              & 280      & 2.85                    \\
24  & 38              & 300      & 1.79                    & 56  & 90              & 300      & 2.76                    \\
25  & 50              & 80       & 3.98                    & 57  & 100             & 80       & 5.63                    \\
26  & 50              & 100      & 3.56                    & 58  & 100             & 100      & 5.03                    \\
27  & 50              & 150      & 2.91                    & 59  & 100             & 150      & 4.11                    \\
28  & 50              & 180      & 2.65                    & 60  & 100             & 180      & 3.75                    \\
29  & 50              & 220      & 2.40                    & 61  & 100             & 220      & 3.39                    \\
30  & 50              & 260      & 2.21                    & 62  & 100             & 260      & 3.12                    \\
31  & 50              & 280      & 2.13                    & 63  & 100             & 280      & 3.01                    \\
32  & 50              & 300      & 2.05                    & 64  & 100             & 300      & 2.91\\                   
	\hline
	\end{tabular}}

\end{table}

Three indicators are utilized in this study to evaluate the MC-GRU: Mean Square Error (MSE), Mean Absolute Error (MAE), and Coefficient of Determination (\(R^2\)), as expressed in Eq. \ref{eqmse}-\ref{eqr2}. MSE and MAE provide direct quantification of the disparities between predicted and actual values, with lower values indicating superior predictive performance. Conversely, \(R^2\) offers an overall evaluation of the model fit, with values closer to 1 denoting better fitting.

\begin{flalign}
    \label{eqmse}
&MSE = \frac{1}{n} \sum_{i=1}^{n} (y_i - \hat{y}_i)^2 &
\\
    \label{eqmae}
&MAE = \frac{1}{n} \sum_{i=1}^{n} |y_i - \hat{y}_i| &
\\
    \label{eqr2}
&R^2 = 1 - \frac{\sum_{i=1}^{n} (y_i - \hat{y}_i)^2}{\sum_{i=1}^{n} (y_i - \bar{y})^2} &
\end{flalign}
where, \( n \) is the sample size, \( y_i \) is the actual value of the \( i \)th sample, \( \hat{y}_i \) is the predicted value for the \( i \)th sample, and \( \bar{y} \) is the mean of all actual values.

\begin{figure}
    \centering
    \includegraphics[width=1.0\linewidth]{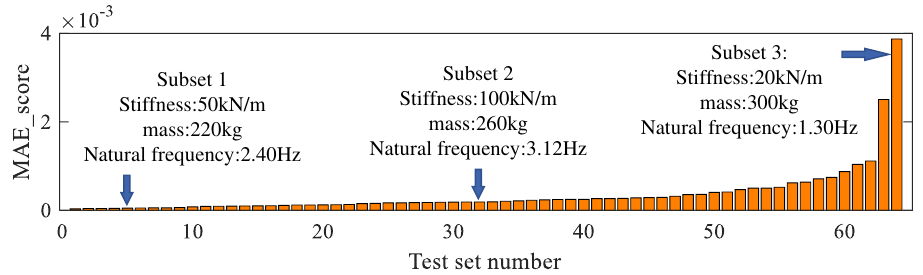}
    \caption{The MAE distribution across subsets with different structural information.}
    \label{fig:linearMAE}
\end{figure}

The MC-GRU achieves an MSE of \( 8.46\times{10}^{-7}\), an MAE of \(3.59\times{10}^{-4}\), and an \(R^2\) score of 0.9436 on the entire test set, indicating strong generalization ability and high prediction accuracy for unseen structures. To further evaluate its generalization across different structures, the test set is divided into 64 subsets corresponding to different structural characteristics shown in Table \ref{table2}, and the MAE for each subset is calculated and ranked in ascending order, as shown in Figure \ref{fig:linearMAE}. Additionally, three representative subsets are selected (highlighted with arrows in Figure \ref{fig:linearMAE}) to clearly show the model generalizable performance across different structures. 

\begin{figure}[!t]
    \centering
    \includegraphics[width=1\linewidth]{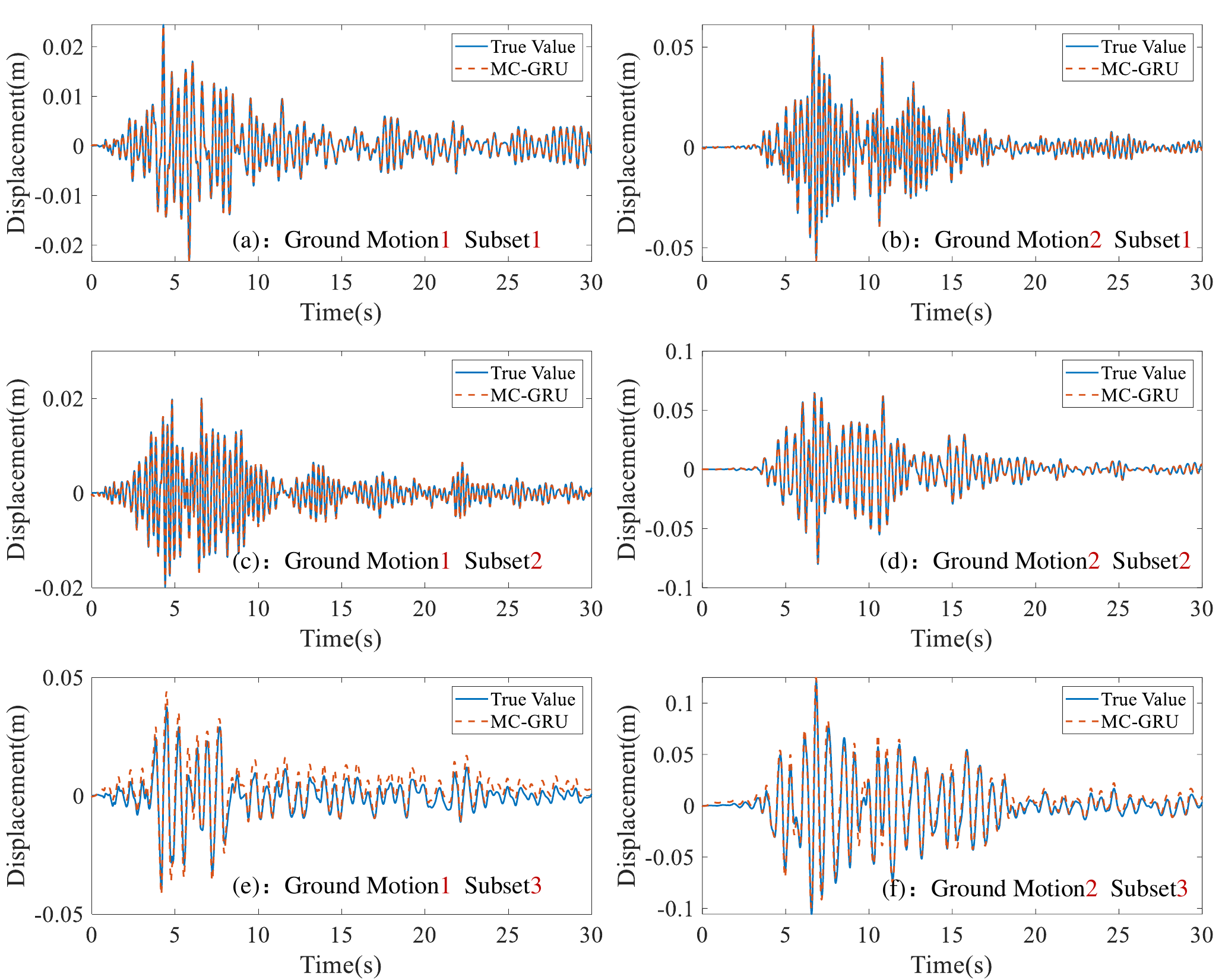}
    \caption{The response prediction results for three representative subsets.}
    \label{fig:Linear test}
\end{figure}

Figure \ref{fig:Linear test} shows the time history response prediction of three representative subsets with different levels of accuracy. Subset 1 features a structural stiffness of 50 kN/m and a mass of 220 kg, and the performance indicators are an MSE of \(7.92\times{10}^{-9}\), an MAE of \(5.01\times{10}^{-5}\), and an \(R^2\) score of 0.9997. These metrics rank 7th, 5th, and 5th out of 64 in the test dataset, respectively. The predicted structural responses for Subset 1 are shown in Figures \ref{fig:Linear test}(a) and (b).
Subset 2 has a structural stiffness of 100 kN/m and a mass of 260 kg, and the corresponding performance indicators are an MSE of \(9.90\times{10}^{-8}\), an MAE of \(1.88\times{10}^{-4}\), and an \(R^2\) score of 0.9951, ranking 37th, 32nd, and 34th, respectively.
The predicted structural responses for Subset 2 are shown in Figures \ref{fig:Linear test}(c) and (d) under two example earthquakes. For both Subset 1 \& 2, the predicted responses closely align with the ground truth, indicating a high level of accuracy. For Subset 3, the structural stiffness and mass are 20 kN/m and 300 kg, respectively. The performance indicators include an MSE of \(2.53\times{10}^{-5}\), an MAE of 0.0039, and an \(R^2\) score of 0.7858. These rank 64th, 64th, and 59th out of 64, respectively, making Subset 3 the worst-performing subset in terms of prediction accuracy. Figures \ref{fig:Linear test}(e) and (f) show the predicted structural responses. The worse performance of Subset 3 could probably be attributed to its low stiffness and high mass, leading to a lower natural frequency and a greater difference from the training set compared to other subsets. It is worth noting the displacement responses depicted in Figures \ref{fig:Linear test}(a), (c), and (e) originate from the same ground motion, as do the displacement responses in Figures \ref{fig:Linear test}(b), (d), and (f). These figures highlight the capability of the MC-GRU model in accurately capturing the effects of structural information discrepancies on displacement predictions. Even for Subnet 3 with a relative large MAE, a good agreement can be still achieved between the prediction and the ground truth, indicating great generalizability of the proposed approach across varying structures.

\subsection{Case 2: Bouc-Wen hysteresis model}\label{sec3.2}
The proposed MC-GRU network is further validated through a nonlinear hysteresis system to test its prediction performance and generalizability when structures experience large deformations under earthquakes especially for those entering plastic stage with residual deformations. The Bouc-Wen hysteresis model \cite{41:SUN201359,42:doi:10.1061/JMCEA3.0002106} is selected as an example case study which is governed by the following equations:

\begin{flalign}
    \label{eq11}
&m\ddot{x}\left(t\right)+c\dot{x}\left(t\right)+f\left(t\right)=-m{\ddot{x}}_g\left(t\right)&
\end{flalign}
where \(x\), \(\dot{x}\), and \(\ddot{x}\) are the vectors of structural displacement, velocity, and acceleration respectively; m and c are the mass and damping coefficient; \({\ddot{x}}_g\) is the ground motion acceleration; and f is the nonlinear hysteretic restoring force vector which can be obtained from \cite{sato1998adaptive}:
\begin{flalign}
    \label{eq12}
&\dot{f}=k\dot{x}-\alpha\left|\dot{x}\right|\left|f\right|^{n-1}f-\beta \dot{x}\left|f\right|^n&
\end{flalign}
where \(k\) is the stiffness; and \(\alpha\) , \(\beta\), and \(n\) are the nonlinear parameters of the Bouc-Wen model (\(\alpha\)=1,  \(\beta\)=2, \(n\)=3 in this study).

Same input information as in Case 1 is utilized for both ground motion excitation and structural information,  yielding a training dataset with 2500 samples, a validation dataset with 5120 samples, and a testing dataset with 1920  samples. Each sample includes ground motion and structural information as inputs, with structural displacement responses as outputs.

\begin{figure}
    \centering
    \includegraphics[width=1.0\linewidth]{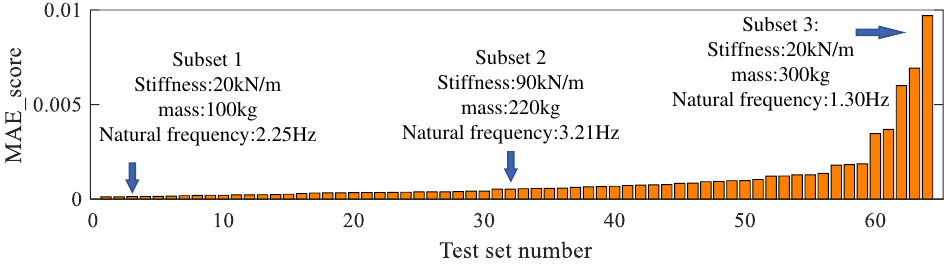}
    \caption{The MAE distribution across subsets with different structural information.}
    \label{fig-MAE2}
\end{figure}

\begin{figure}[b!]
    \centering
    \includegraphics[width=1\linewidth]{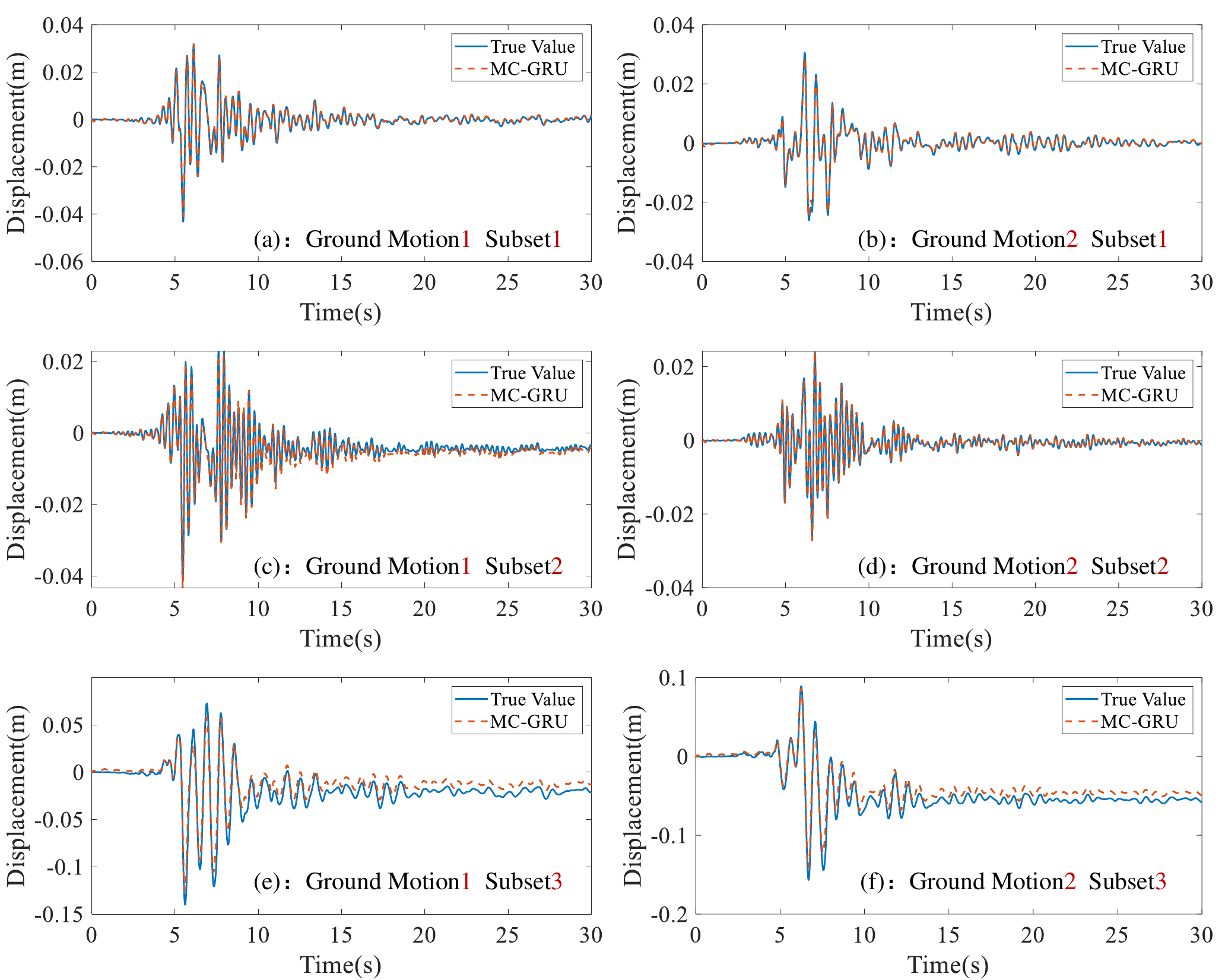}
    \caption{The response prediction results for three representative subsets with Bouc-Wen hysterestic model.}
    \label{fig:BW test}
\end{figure}

The MC-GRU achieved an MSE of \( 3.36 \times 10^{-5} \), an MAE of 0.001, and an \( R^2 \) score of 0.8835 on the entire testing dataset. Overall, the proposed approach maintains sufficient accuracy and generalizability on different unseen structures experiencing plastic deformations. Similarly, the testing dataset is divided into 64 subsets based on varying structural characteristics as shown in Figure \ref{fig-MAE2} with an ascending order of MAE. Three representative subsets with different levels of accuracy are selected to further illustrate the generalizability across varying structures by presenting the details of time-history response prediction as shown in Figure \ref{fig:BW test}. Note that same ground motion input is considered to show the influence of structural characteristics on dynamic responses (e.g., Ground Motion 1 for Figures \ref{fig:BW test} (a), (c), (e) and Ground Motion 2 for Figures \ref{fig:BW test} (b), (d), and (f)).   

The Subset 1 in Figure \ref{fig-MAE2} has a structural stiffness of 20 kN/m and a mass of 100 kg, with an MSE of \(3.25\times{10}^{-8}\), an MAE of \(1.37\times{10}^{-4}\), and an \(R^2\) score of 0.9964. These metrics rank 2nd, 3rd, and 4th out of 64, respectively, across the entire testing dataset. Figures \ref{fig:BW test}(a) and (b) show the predicted structural responses for Subset 1 under two example earthquakes. Good matching is observed with no significant residual displacements. The Subset 2 has a structural stiffness of 90 kN/m and a mass of 220 kg, achieving an MSE of \(9.59\times{10}^{-7}\), an MAE of \(5.35\times{10}^{-4}\), and an \(R^2\) score of 0.9697, with a ranking of 39th, 32nd, and 35th, respectively. The predicted structural responses for Subset 2 are shown in Figures \ref{fig:BW test}(c) and (d). Residual deformation is observed in Figure \ref{fig:BW test}(c) and our model can accurately predict it. The Subset 3 has a structural stiffness of 20 kN/m and a mass of 300 kg, achieving an MSE of \(8.16\times{10}^{-4}\), an MAE of 0.0097, and an \(R^2\) score of 0.7166, ranking 64th, 64th, and 59th, respectively, across the entire testing dataset. Figures \ref{fig:BW test}(e) and (f) show the predicted structural responses for Subset 3 with significant residual displacements observed. Even under such circumstance, the proposed MC-GRU approach is capable of accurately predicting the dynamic responses, demonstrating favorable generalizability across varying structures.

\begin{figure}
    \centering
    \includegraphics[width=1\linewidth]{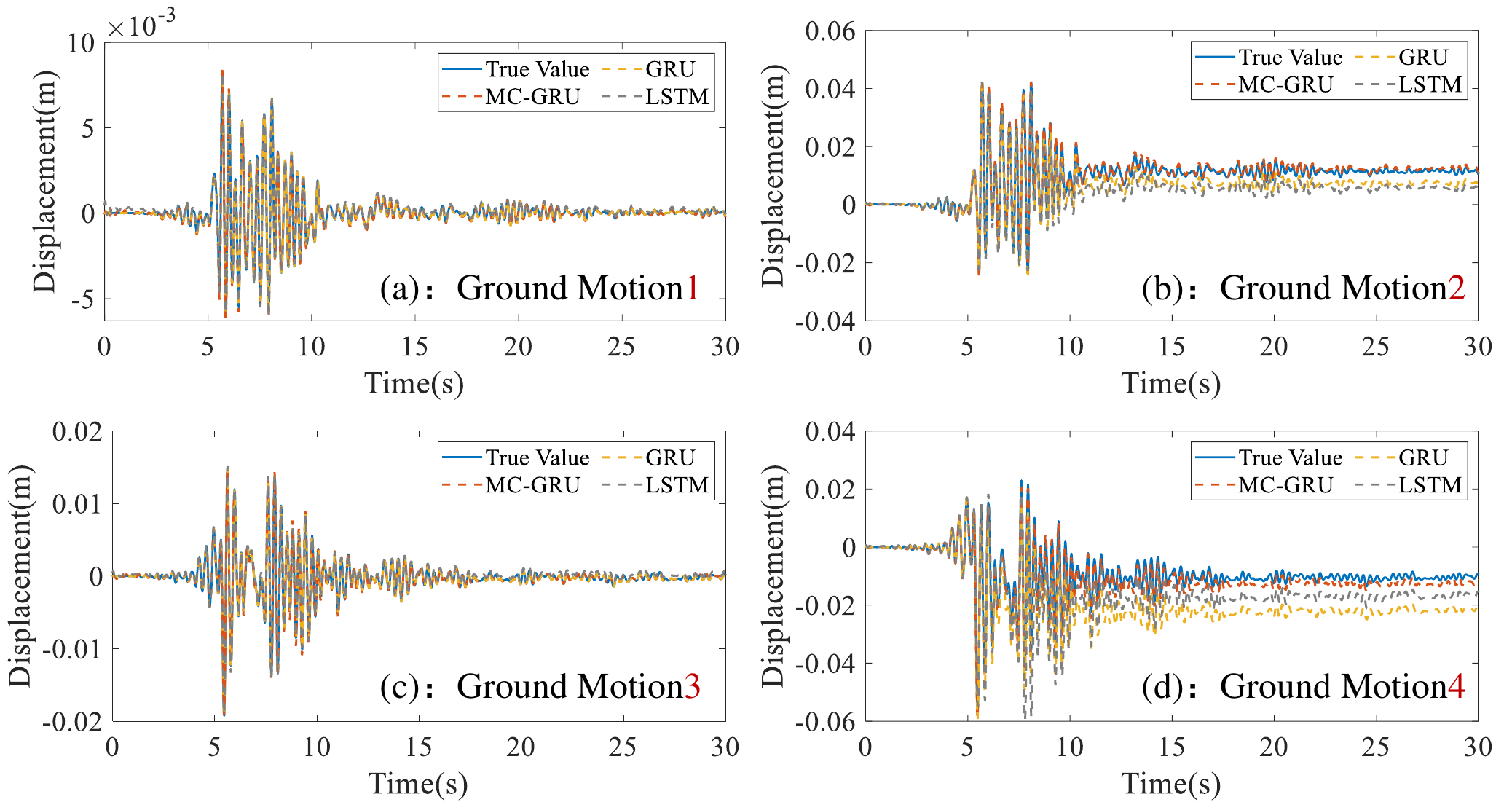}
    \caption{The Bouc-Wen hysterestic model response prediction results of LSTM, GRU, and MC-GRU.}
    \label{fig:BW test2}
\end{figure}

Additionally, the MC-GRU is compared to classical LSTM and GRU networks to further highlight the performance and generalizability across different structures. It is worth mentioning that both LSTM and GRU are incapable of generalizing on varying structures. Therefore, the LSTM and GRU models are trained for a fixed structure with a stiffness of 90 kN/m and a mass of 220 kg (same structure of Subset 2), excited by same ground motion inputs as the MC-GRU network. Table \ref{BW:MSE MAE R2} presents the MSE, MAE, and \( R^2 \) values of three networks on entire testing set for Subset 2 structure. It is clear that the MC-GRU outperforms both LSTM and GRU on all evaluation metrics. Same conclusion can be obtained from Figure \ref{fig:BW test2}, which shows the time history responses under four example earthquakes. The MC-GRU is capable of accurately predicting the peaks and residuals even though the structure information is not given in training dataset. It further highlights the generalizability and robustness of the proposed MC-GRU approach on capturing dynamic characteristics of varying structures.

\begin{table}[t!]
	\caption{The MSE, MAE, and $\text{R}^2$ on the test set, for the LSTM, GRU, and MC-GRU networks.} \vspace{6pt}
	\label{BW:MSE MAE R2}
	\footnotesize
	\centering
	\setlength{\tabcolsep}{5mm}{
	\begin{tabular}{ccccccc} 
	\hline 
Algorithm & MSE & MAE & $\text{R}^2$ 
\\ \hline
MC-GRU   & \(9.59\times{10}^{-7}\)       & \(5.35\times{10}^{-4}\)                     & 0.9697    \\
GRU   & \( 5.58\times{10}^{-6}\)       & \(9.55\times{10}^{-4}\)                    & 0.8949       \\
LSTM   & \( 1.28\times{10}^{-5}\)       & \(1.80\times{10}^{-3}\)                     & 0.7192 \\
 \hline
	\end{tabular}}
\end{table}

The correlation index (\(CI\)) is also used to evaluate the prediction performance of these models, which is defined as follows:
\begin{equation}
CI\left(Y_i, Y_j\right)=\frac{Cov\left(Y_i, Y_j\right)}{\sqrt{Var\left(Y_i\right)Var\left(Y_j\right)}}
\end{equation}
where the term \(CI\) denotes the correlation index between the sequences \(Y_i\) and  \(Y_j\). The operators \(Cov()\) and \(Var()\) represent the covariance and variance, respectively.

\begin{figure}[t!]
    \centering
    \includegraphics[width=1\linewidth]{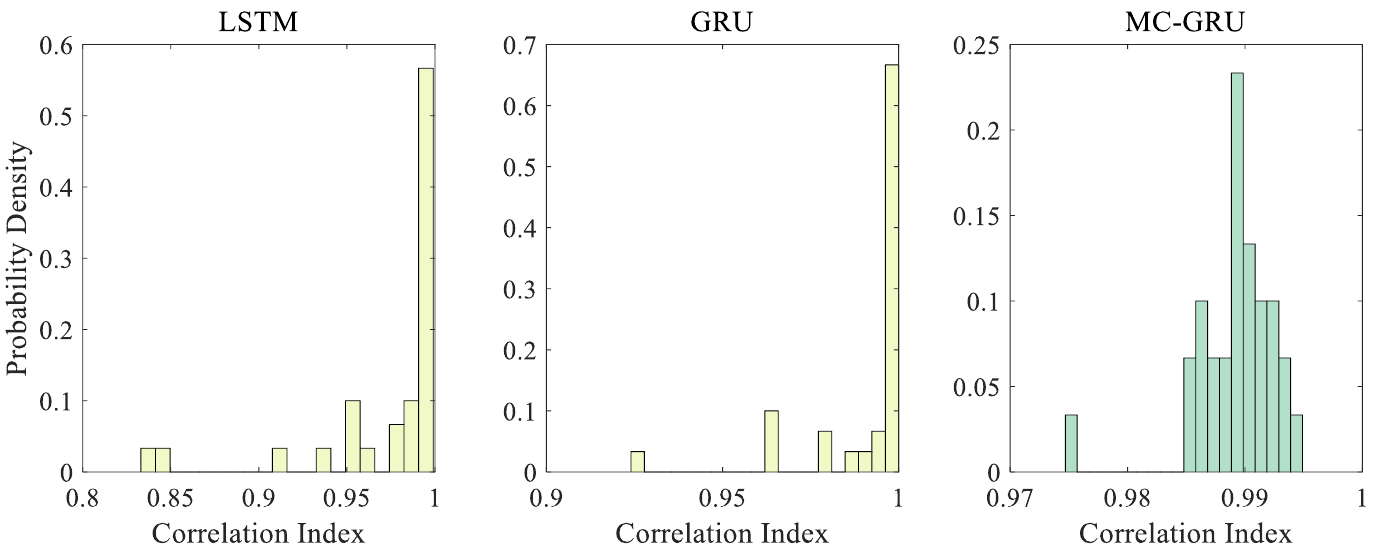}
    \caption{The probability distribution results of CI for LSTM, GRU, and MC-GRU.}
    \label{fig:BW ci}
\end{figure}

Figure \ref{fig:BW ci} compares the probability distribution of the CI for the proposed MC-GRU with classical LSTM and GRU. Both LSTM and GRU achieve an acceptable CI distribution. However, the model is trained for the specific structure with no generalizability on other structures. The MC-GRU achieves a more concentrated  and higher CI distribution greater than 0.97, outperforming baseline GRU and LSTM, even though the MC-GRU is not trained on this specific structure. It is credited to the specific design of MC-GRU network by incorporating structural parameters to enrich the nonlinear dynamic characteristics and enhance the learning and generalizability on varying structures. 

\subsection{Case 3: Reinforced concrete column}\label{sec3.3}

\begin{figure}[thbp!]
    \centering
    \includegraphics[width=1.0\linewidth]{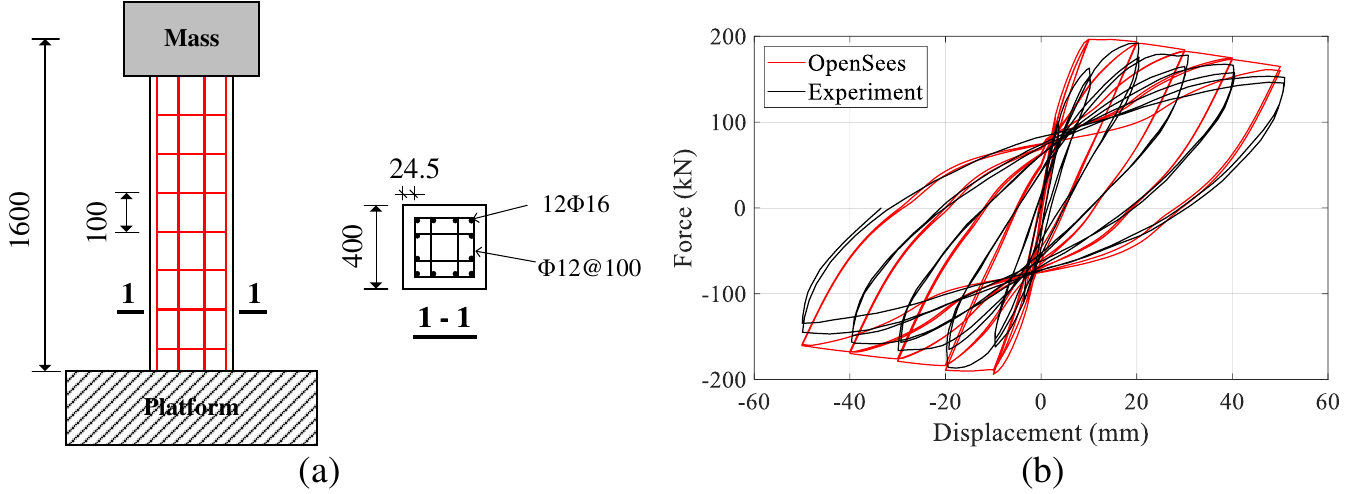}
    \caption{(a) The dimensions and sectional of the RC column(unit:mm); (b) The hysteresis curves of the RC-column.}
    \label{fig:Column}
 \end{figure}

The proposed MC-GRU network is further validated experimentally based on Ang et al. 1981, No. 3 \cite{ang1981ductility} (rectangular) from the PEER Structural Performance Database \cite{PEER}. The dimensions and sectional layout of the RC column are provided in Figure \ref{fig:Column} (a). The RC column has a height of 1600 mm and a square cross-section with a side length of 400 mm. Twelve longitudinal reinforcement bars, each with a diameter of 16 mm, are used. The stirrups, 12 mm in diameter, are spaced at 100 mm intervals along the column height. The yield stress and ultimate tensile strength of the longitudinal reinforcement are 427 MPa and 670 MPa, respectively. And the concrete has a compressive strength of 23.6 MPa. To achieve an axial compression ratio of 0.38, a load of 1435 kN is applied to the top of the column. The finite element model of the RC column is first developed using OpenSees and validated by comparing to experimental results. Figure \ref{fig:Column} (b) presents the hysteresis curves with cyclic lateral forces applied at the top of the column. The backbone curves of the simulated hysteresis curves are found to closely align with experimental results, confirming the accuracy of the finite element model. 

\begin{figure}
    \centering
    \includegraphics[width=1\linewidth]{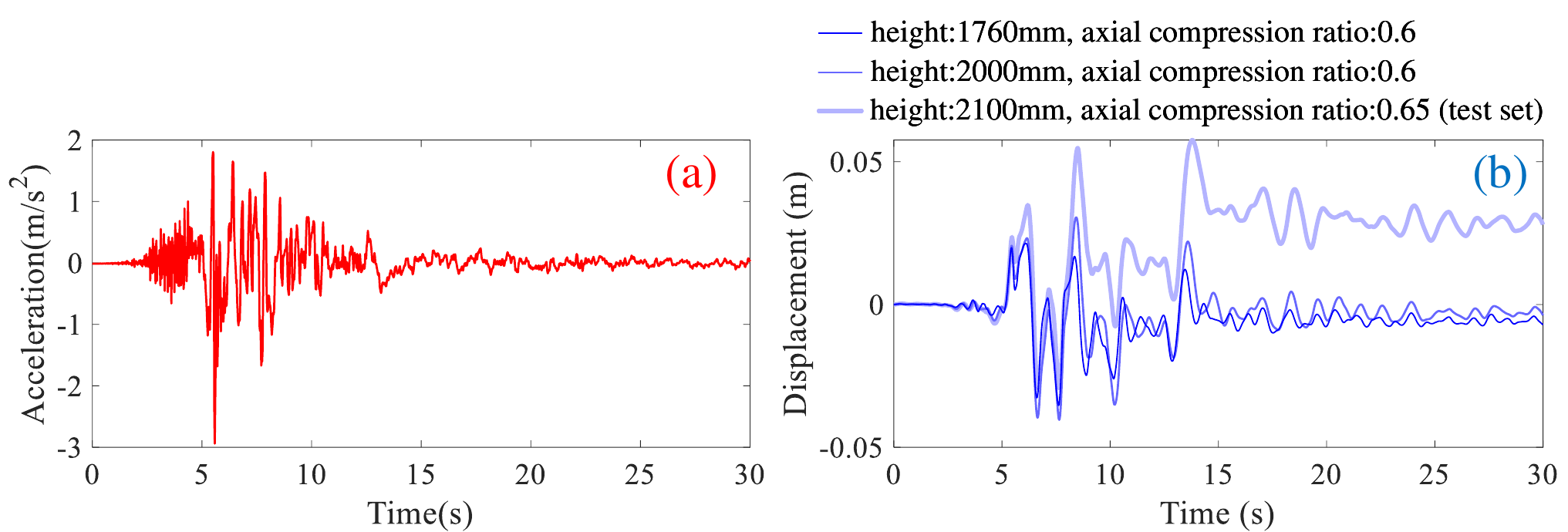}
    \caption{The displacement response of different columns under the same ground motion: (a)ground motion acceleration. (b)displacement response of different columns(No.29, No.30 in Table \ref{table-RC-column-train}) and No.2 in Table \ref{table-RC-column-test}}
    \label{fig:ag_x}
\end{figure}

To evaluate the generalizability performance of the proposed MC-GRU on varying structure parameters, a synthetic dataset is established through changing the height and axial compression ratio of the RC column. The adjustment of parameters stays within the reasonable range specified by the Chinese seismic code \cite{China}. Details of the structural information are summarized in Table \ref{table-RC-column-train}, Table \ref{table-RC-column-validate}, and Table \ref{table-RC-column-test} for training, validation, and testing, respectively. The training dataset consists of 30 columns, with column heights ranging from 800 mm to 2000 mm and axial compression ratios from 0.2 to 0.6. The natural frequencies are distributed between 1.35 Hz and 10.41 Hz, as shown in Table \ref{table-RC-column-train}. 
The validation dataset consists of 3 columns, with column heights ranging from 1500 mm to 1900 mm and axial compression ratios from 0.25 to 0.55. The natural frequencies range from 2.06 Hz to 2.71 Hz, as shown in Table \ref{table-RC-column-validate}. The testing dataset consists of 5 columns, with column heights ranging from 1000 mm to 2100 mm and axial compression ratios from 0.1 to 0.65. The natural frequencies are from 1.18 Hz to 4.71 Hz, presented in Table \ref{table-RC-column-test}. It is important to note that the structural information used for testing is totally unseen, including different column heights and axial compression ratios out of the range of the training dataset. To clearly show the difference of varying structures, the effect of varying structural parameters on the displacement time history response of the columns is shown in Figure \ref{fig:ag_x}. It includes one sample from the test dataset alongside two training samples most similar to it. It can be clearly seen the difference of displacement responses given different parameters under same seismic excitation. The proposed MC-GRU is expected to learn structural dynamic characteristics and generalize to different unseen structures.

\begin{table}[t!]
	\caption{The RC-column structural information in training datasets.} \vspace{6pt}
	\label{table-RC-column-train}
	\footnotesize
	\centering
	\setlength{\tabcolsep}{5mm}{
	\begin{tabular}{ccccccc} 
	\hline 
No. & \begin{tabular}[c]{@{}c@{}} Height \\ (mm)\end{tabular} & Axial compression ratio & \begin{tabular}[c]{@{}c@{}}Mass \\ (t)\end{tabular} & \begin{tabular}[c]{@{}c@{}}Stiffness \\ (kN/m)\end{tabular} & \begin{tabular}[c]{@{}c@{}} Natural frequency\\(Hz)\end{tabular}               \\ \hline
1   & 800        & 0.2                     & 77.06       & \( 3.30\times{10}^{5}\) & 10.41 \\
2   & 1040       & 0.2                     & 77.06       & \( 1.50\times{10}^{5}\) & 7.01 \\
3   & 1280       & 0.2                     & 77.06 & \( 8.00\times{10}^{4}\) & 5.13 \\
4   & 1520       & 0.2                     & 77.06 & \( 4.76\times{10}^{4}\) & 3.96 \\
5   & 1760       & 0.2                     & 77.06 & \( 3.01\times{10}^{4}\) & 3.17 \\
6   & 2000       & 0.2                     & 77.0 & \( 2.07\times{10}^{4}\) & 2.61  \\
7   & 800        & 0.3                     & 115.59 & \( 3.17\times{10}^{5}\) & 8.34 \\
8   & 1040       & 0.3                     & 115.59 &\( 1.44\times{10}^{5}\) & 5.64 \\
9   & 1280       & 0.3                     & 115.59 & \( 7.68\times{10}^{4}\) & 4.10 \\
10  & 1520       & 0.3                     & 115.59 & \( 4.56\times{10}^{4}\) & 3.16 \\
11  & 1760       & 0.3                     & 115.59 & \( 2.92\times{10}^{4}\) & 2.53 \\
12  & 2000       & 0.3                     & 115.59 & \( 1.98\times{10}^{4}\) & 2.08 \\
13  & 800        & 0.4                     & 154.12  & \( 3.04\times{10}^{5}\) & 7.07 \\
14  & 1040       & 0.4                     & 154.12  & \( 1.38\times{10}^{5}\) & 4.76 \\
15  & 1280       & 0.4                     & 154.12  & \( 7.35\times{10}^{4}\) & 3.47 \\
16  & 1520       & 0.4                     & 154.12 & \( 4.35\times{10}^{4}\) & 2.67 \\
17  & 1760       & 0.4                     & 154.12  & \( 2.78\times{10}^{4}\) & 2.14 \\
18  & 2000       & 0.4                     & 154.12  & \( 1.87\times{10}^{4}\) & 1.76 \\
19  & 800        & 0.5                     & 192.65 & \( 2.91\times{10}^{5}\)  & 6.19 \\
20  & 1040       & 0.5                     & 192.65 & \( 1.32\times{10}^{5}\) & 4.16 \\
21  & 1280       & 0.5                     & 192.65 & \( 7.00\times{10}^{4}\)  & 3.03 \\
22  & 1520       & 0.5                     & 192.65 & \( 4.14\times{10}^{4}\) & 2.33 \\
23  & 1760       & 0.5                     & 192.65 & \( 2.63\times{10}^{4}\) & 1.86 \\
24  & 2000       & 0.5                     & 192.65 & \( 1.77\times{10}^{4}\) & 1.53 \\
25  & 800        & 0.6                     & 231.18 & \( 2.77\times{10}^{5}\) & 5.51 \\
26  & 1040       & 0.6                     & 231.18 & \( 1.25\times{10}^{5}\)  & 3.70 \\
27  & 1280       & 0.6                     & 231.18 & \( 6.64\times{10}^{4}\) & 2.70 \\
28  & 1520       & 0.6                     & 231.18 & \( 3.91\times{10}^{4}\) & 2.07 \\
29  & 1760       & 0.6                     & 231.18 & \(2.48\times{10}^{4}\) & 1.65 \\
30  & 2000       & 0.6                     & 231.18 & \( 1.66\times{10}^{4}\) & 1.35 \\ \hline
	\end{tabular}}
\end{table}

\begin{table}[t!]
	\caption{The RC-column Structural information in validation datasets.} \vspace{6pt}
	\label{table-RC-column-validate}
	\footnotesize
	\centering
	\setlength{\tabcolsep}{5mm}{
	\begin{tabular}{ccccccc} 
	\hline 
No. & \begin{tabular}[c]{@{}c@{}} Height \\ (mm)\end{tabular} & Axial compression ratio & \begin{tabular}[c]{@{}c@{}}Mass \\ (t)\end{tabular} & \begin{tabular}[c]{@{}c@{}}Stiffness \\ (kN/m)\end{tabular} & \begin{tabular}[c]{@{}c@{}} Natural frequency\\(Hz)\end{tabular}               \\ \hline
1   & 1800        & 0.25                     & 96.33       & \( 2.79\times{10}^{4}\) & 2.71 \\
2   & 1500       & 0.55                     & 211.92       & \( 4.19\times{10}^{4}\) & 2.24 \\
3   & 1900       & 0.35                     & 134.86 & \( 2.25\times{10}^{4}\) & 2.06 \\
 \hline
	\end{tabular}}
\end{table}

\begin{table}[t!]
	\caption{The RC-column Structural information in testing datasets.} \vspace{6pt}
	\label{table-RC-column-test}
	\footnotesize
	\centering
	\setlength{\tabcolsep}{5mm}{
	\begin{tabular}{ccccccc} 
	\hline 
No. & \begin{tabular}[c]{@{}c@{}} Height \\ (mm)\end{tabular} & Axial compression ratio & \begin{tabular}[c]{@{}c@{}}Mass \\ (t)\end{tabular} & \begin{tabular}[c]{@{}c@{}}Stiffness \\ (kN/m)\end{tabular} & \begin{tabular}[c]{@{}c@{}} Natural frequency\\(Hz)\end{tabular}               \\ \hline
1   & 1600        & 0.38                     & 146.42       & \( 3.76\times{10}^{4}\) & 2.55 \\
2   & 2100       & 0.65                     & 250.45       & \( 1.37\times{10}^{4}\) & 1.18 \\
3   & 2100       & 0.15                     & 57.80 & \( 1.83\times{10}^{4}\) & 2.83 \\
4   & 2100      & 0.1                     & 38.53 & \( 1.87\times{10}^{4}\) & 3.51 \\
5   & 1000       & 0.45                     & 173.39 & \( 1.51\times{10}^{5}\) & 4.71 \\
 \hline
	\end{tabular}}
\end{table}

\begin{figure}[t!]
    \centering
    \includegraphics[width=0.5\linewidth]{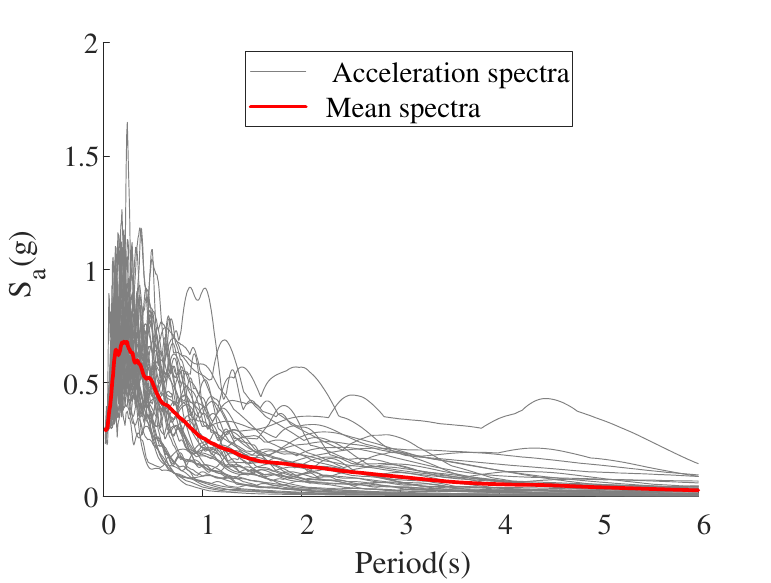}
    \caption{The acceleration response spectra.}
    \label{fig:spectra}
\end{figure}

Same ground motion records used in previous cases are selected as the input excitation, including 20 records to generate training data, 16 records for validation, and 6 records for testing. The amplitude of ground motion data is also scaled to achieve peak a ground accelerations (PGA) of 0.1g, 0.2g, 0.3g, 0.4g, and 0.5g to ensure the columns entering different damage levels. Figure \ref{fig:spectra} shows the acceleration spectra of the scaled ground motion records.

\begin{figure}
    \centering
    \includegraphics[width=1\linewidth]{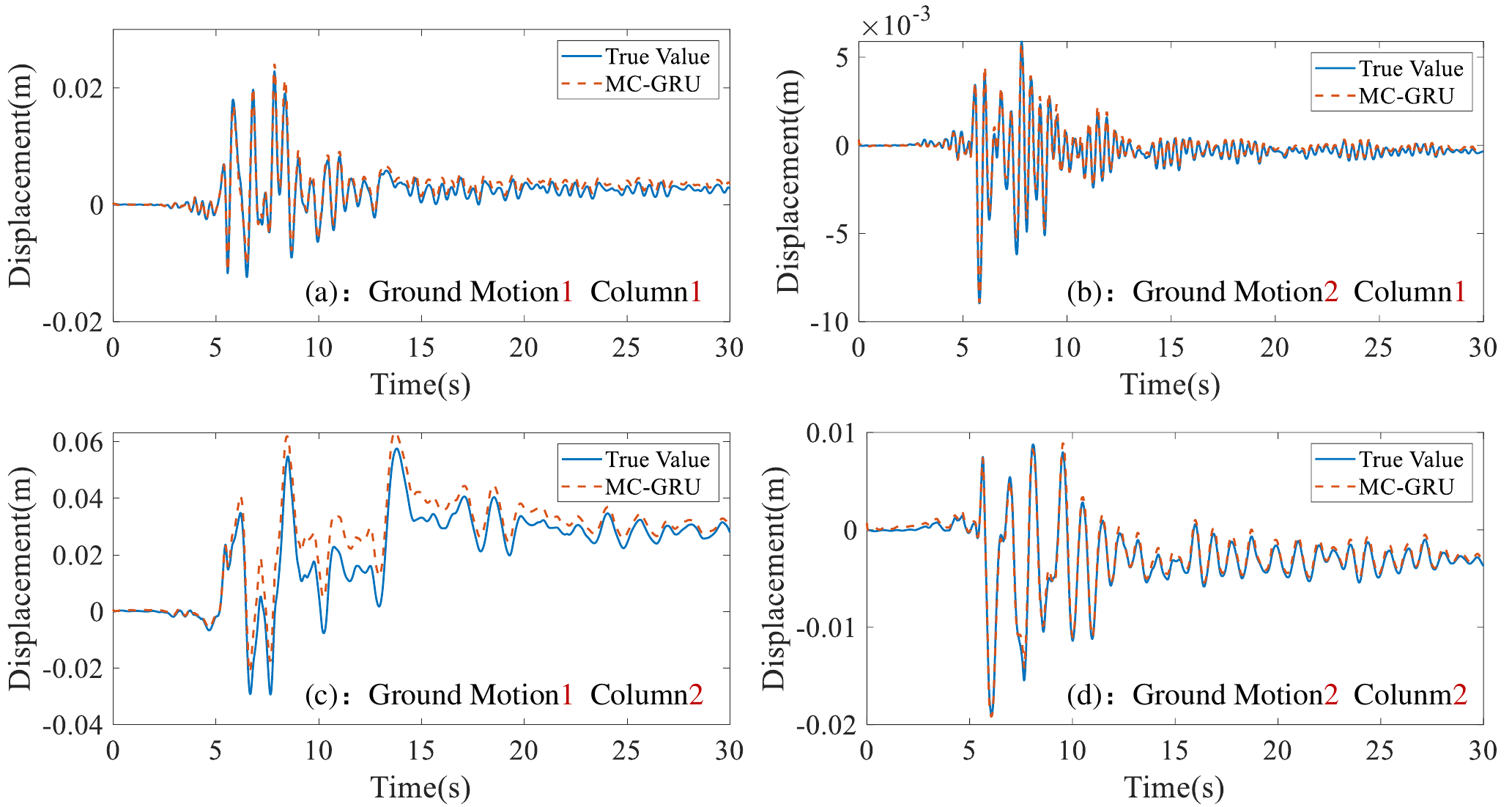}
    \caption{The displacement response predictions of column 1 and column 2 in the testing set.}
    \label{fig:Column-test}
\end{figure}

The generalization performance of the MC-GRU is evaluated across all testing dataset in terms of MSE, MAE, and \(R^2\) indexes, as shown in Table \ref{table-RC-column-lstmgru}. The column 1 (with a height of 1600 mm and an axial compression ratio of 0.38) and column 2 (with a height of 2100 mm and an axial compression ratio of 0.65) exhibit the best and worst performance, respectively. Therefore, column 1 and column 2 of the testing dataset are selected as the representative examples. Note that column 1 is the testing column shown in Figure \ref{fig:Column}. Figure \ref{fig:Column-test} shows the predicted time history of displacements of column 1 \& 2 under two example earthquakes. By comparing Figures \ref{fig:Column-test} (a) \& (c) or Figures \ref{fig:Column-test} (b) \& (d), we can clearly see the difference of two columns reacted under same excitation. It can be seen that the predicted displacement responses for both column 1 and column 2 closely match the ground truth. Since column 2 has lower stiffness and higher mass than column 1, it experiences larger deformation with a residual displacement around 3 cm (as shown in Figure \ref{fig:Column-test} (c)). It is worth mentioning that both the height and axial compression ratio of column 2 are out of the range of the parameters in training dataset. Even under such circumstance, the MC-GRU approach can accurately predict the responses with only minor deviations on residuals. It clearly demonstrates the generalizable capability of MC-GRU on varying unseen structures.

\begin{table}[t!]
	\caption{The MSE,MAE and $\text{R}^2$ of MC-GRU in testing datasets.} \vspace{6pt}
	\label{table-RC-column-lstmgru}
	\footnotesize
	\centering
	\setlength{\tabcolsep}{5mm}{
	\begin{tabular}{ccccccc} 
	\hline 
No. & \begin{tabular}[c]{@{}c@{}} Height \\ (mm)\end{tabular} & Axial compression ratio & MSE & MAE & $\text{R}^2$               \\ \hline
1   & 1600 & 0.38 & \( 8.43\times{10}^{-7}\)   & \( 4.08\times{10}^{-4}\) & 0.9679 \\
2   & 2100   & 0.65       & \( 3.39\times{10}^{-4}\)       & \( 7.40\times{10}^{-3}\) & 0.3815 \\
3   & 2100  & 0.15        & \(2.03\times{10}^{-7}\) & \(3.11\times{10}^{-4}\) & 0.9343 \\
4   & 2100 & 0.1   & \(4.96\times{10}^{-7}\) & \( 4.10\times{10}^{-4}\) & 0.7614 \\
5   & 1000    & 0.45       & \(1.56\times{10}^{-7}\) & \( 2.18\times{10}^{-4}\) & 0.8547 \\
 \hline
	\end{tabular}}
\end{table}

\begin{figure}
    \centering
    \includegraphics[width=1\linewidth]{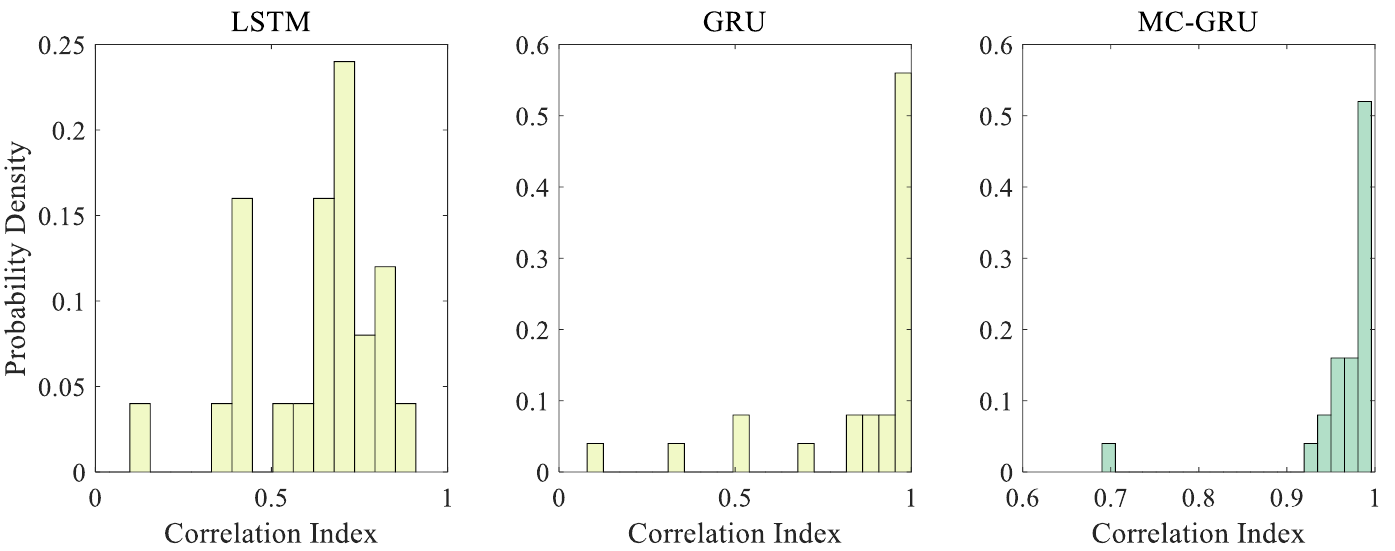}
    \caption{The probability distribution results of CI for LSTM, GRU, and MC-GRU.}
    \label{fig:Cloumn ci}
\end{figure}

\begin{figure}[H]
    \centering
    \includegraphics[width=1\linewidth]{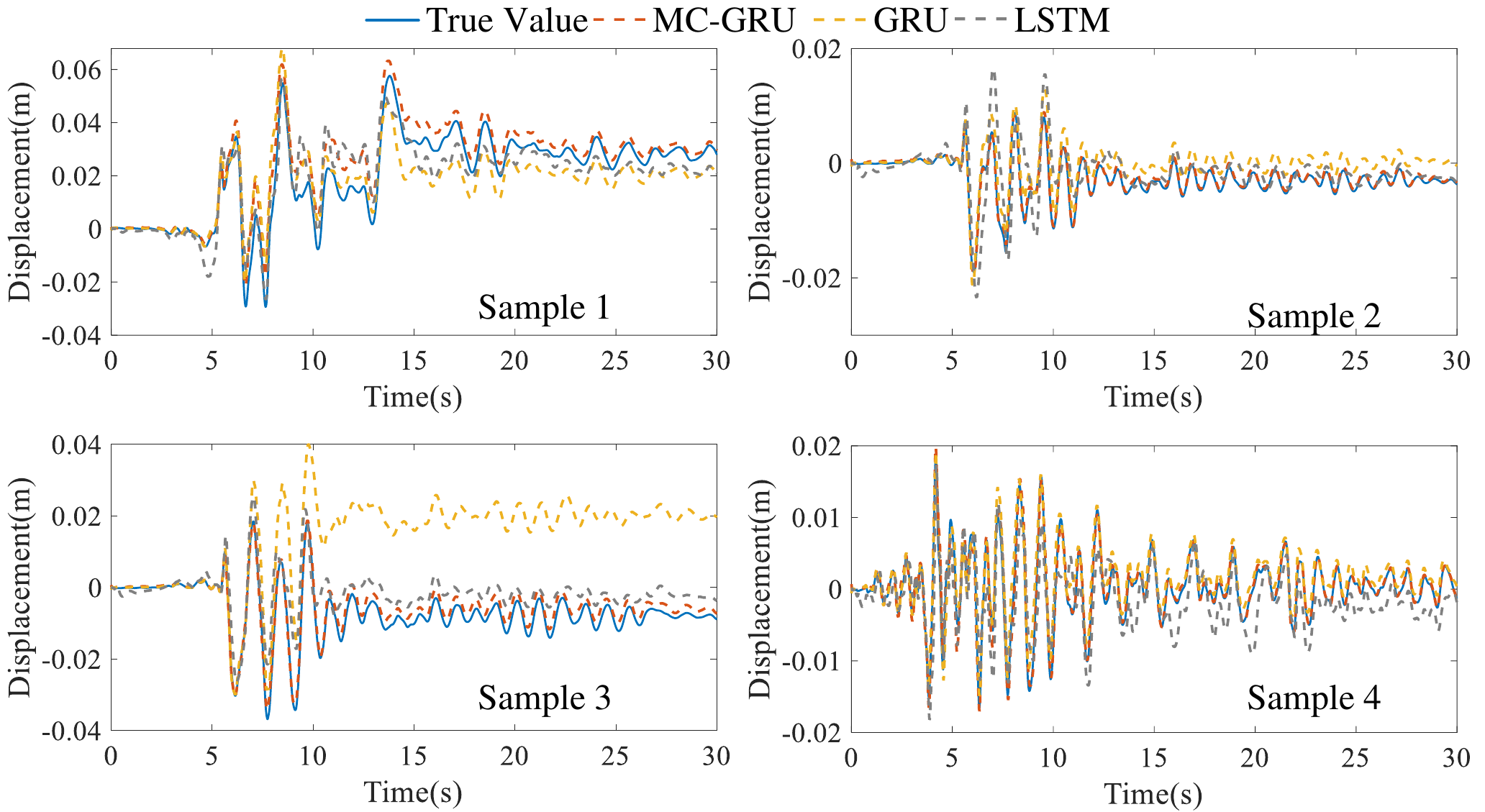}
    \caption{The nonlinear response prediction results of the proposed MC-GRU on diverse structures compared to classical LSTM and GRU.}
    \label{fig:Column-test-grulstm}
\end{figure}

To further highlight the breakthrough performance of the proposed approach on generalizability, a comparison against classical LSTM and GRU networks is conducted. Similar to the previous case study, the LSTM and GRU models are trained independently for column 2. The overall performance is presented in Table \ref{table-RC-M,M,R} and Figure \ref{fig:Cloumn ci}. Table \ref{table-RC-M,M,R} summarizes the MSE, MAE, and \(R^2\) indexes on same testing dataset for the LSTM, GRU, and MC-GRU model. Figure \ref{fig:Cloumn ci} presents the probability distribution of the CI for the LSTM, GRU, and MC-GRU model on all testing dataset. Both LSTM and GRU display discrete distributions with many samples below 0.8. In contrast, the MC-GRU exhibits a sharp and highly concentrated CI distribution near 1. Additionally, the predicted displacement time-history using LSTM, GRU, and MC-GRU are compared and presented in Figure \ref{fig:Column-test-grulstm}. It can be clearly seen that the MC-GRU better matches the ground truth in terms of both peaks and residuals. It clearly shows that the generalizable performance of the MC-GRU on column 2 as unseen structures outperforms the LSTM and GRU models even though they are trained specifically for column 2. This further demonstrates that the proposed MC-GRU well tackles the generalizability issues of existing deep learning approaches on varying structures.

\begin{table}[H]
	\caption{MSE, MAE, and $\text{R}^2$ of LSTM, GRU, and MC-GRU in testing datasets.} \vspace{6pt}
	\label{table-RC-M,M,R}
	\footnotesize
	\centering
	\setlength{\tabcolsep}{5mm}{
	\begin{tabular}{ccccccc} 
	\hline 
Algorithm & MSE & MAE & $\text{R}^2$               \\ \hline
LSTM   & \( 5.94\times{10}^{-4}\) & 0.01 & -1.36    \\
GRU   & \( 4.24\times{10}^{-4}\)  & \(9.40\times{10}^{-3}\)       & -1.08        \\
MC-GRU  & \( 3.39\times{10}^{-4}\)       & \( 7.40\times{10}^{-3}\) & 0.3815 \\
 \hline
	\end{tabular}}
\end{table}

\section{Conclusions}\label{sec_con}
This paper presents a novel MC-GRU network to tackle the  bottleneck of existing AI-based approaches on generalized nonlinear structural response prediction across varying structures. The MC-GRU is specifically designed through a multi-channel input mechanism based on GRU-based gated units to captures the influences of both ground motion and structural information on response prediction, thus achieving favorable generalizability across different structures. The proposed approach is validated through three case studies, including a single-degree-of-freedom linear system, a hysteretic Bouc-Wen model, and a nonlinear reinforced concrete column from experimental testing. 

First, the performance of the MC-GRU network is successfully validated on a simple linear structure. The MC-GRU is generalizable to different linear structures with varying stiffness and mass. Additionally, the MC-GRU is tested on a more complex hysteretic system to validate its generalizability performance on highly nonlinear behaviors across different structures. The results demonstrate that the MC-GRU can accurately predict the nonlinear behaviors including residual deformation for different structures even they enter in plastic zone. The proposed approach is further validated on a reinforced concrete column from experimental testing. The trained MC-GRU successfully predicts the nonlinear displacement responses of unseen columns with different dimensions and weights. Furthermore, the proposed MC-GRU network is compared to classical LSTM and GRU networks to highlight the favorable performance on achieving accuracy and generalizability across varying structures. It demonstrates the MC-GRU outperforms traditional LSTM and GRU networks in capturing nonlinear characteristics of varying structures.

In conclusion, the proposed MC-GRU approach overcomes the major generalizability issues of existing methods, with capability of accurately inferring seismic responses of varying structures. It not only improves the adaptability and robustness of the surrogate model, but also broadens its applicability to a wider range of engineering scenarios. Finally, it is noteworthy that the proposed approach is fundamental in nature which is generalizable to model the dynamics of other types of material and structural systems. The MC-GRU presents a significant advancement in AI-based surrogate modeling and offers significant potential in resilience assessment of urban buildings. 

\section*{Declaration of competing interest}\label{sec_decl}

The authors declare that they have no known competing financial interests or personal relationships that could
have appeared to influence the work reported in this paper.

\section*{Acknowledgement}
This work was financially supported by the National Natural Science Foundation of China (Grant No. 52208466), and the Fundamental Research Funds for the Central Universities. The authors greatly acknowledge the financial support.

\bibliographystyle{elsarticle-num}
\bibliography{refs}

\end{document}